\documentclass{article}

\PassOptionsToPackage{numbers, sort}{natbib}
\usepackage[preprint]{neurips_2022}

\usepackage[utf8]{inputenc} 
\usepackage[T1]{fontenc}    
\usepackage{hyperref}       
\usepackage{url}            
\usepackage{booktabs}       
\usepackage{amsfonts}       
\usepackage{nicefrac}       
\usepackage{microtype}      
\usepackage{xcolor}         
\usepackage{amsmath} 
\usepackage{comment}
\usepackage{graphicx}
\usepackage{algpseudocode}
\usepackage{algorithm}
\usepackage{mathdots}
\usepackage{verbatim}
\usepackage{wrapfig}
\usepackage{float}

\title{Online Transformers with Spiking Neurons for Fast Prosthetic Hand Control}

\author{%
  N. Leroux \\
  Peter Gr\"unberg Institute, \\
  Forschungszentrum J\"ulich, \\
  Aachen, Germany \\
  \texttt{n.leroux@fz-juelich.de}
  \And
  J. Finkbeiner \\
  Peter Gr\"unberg Institute, \\
  Forschungszentrum J\"ulich, \\
  RWTH\\
  Aachen, Germany \\
  \texttt{j.finkbeiner@fz-juelich.de}
  \AND
  E. Neftci \\
  Peter Gr\"unberg Institute, \\
  Forschungszentrum J\"ulich, \\
  RWTH,\\
  Aachen, Germany \\
  \texttt{e.neftci@fz-juelich.de}
}

\begin{document}

\maketitle

\begin{abstract}
    Transformers are state-of-the-art networks for most sequence processing tasks. However, the self-attention mechanism often used in Transformers requires large time windows for each computation step and thus makes them less suitable for online signal processing compared to Recurrent Neural Networks (RNNs). In this paper, instead of the self-attention mechanism, we use a sliding window attention mechanism. We show that this mechanism is more efficient for continuous signals with finite-range dependencies between input and target, and that we can use it to process sequences element-by-element, this making it compatible with online processing. We test our model on a finger position regression dataset (NinaproDB8) with Surface Electromyographic (sEMG) signals measured on the forearm skin to estimate muscle activities. Our approach sets the new state-of-the-art in terms of accuracy on this dataset while requiring only very short time windows of 3.5 ms at each inference step. Moreover, we increase the sparsity of the network using Leaky-Integrate and Fire (LIF) units, a bio-inspired neuron model that activates sparsely in time solely when crossing a threshold. We thus reduce the number of synaptic operations up to a factor of $\times5.3$ without loss of accuracy. Our results hold great promises for accurate and fast online processing of sEMG signals for smooth prosthetic hand control and is a step towards Transformers and Spiking Neural Networks (SNNs) co-integration for energy efficient temporal signal processing.
\end{abstract}

\section{Introduction}
    Surface Electromyography (sEMG) is a technique that senses currents running through muscular fibers’ membrane to measure muscular activity \cite{zheng_surface_2022}. As sEMG signals are triggered by electrical stimuli from the central nervous system, this method is gaining a strong interest as a mean for Human-Machine Interfacing \cite{zheng_surface_2022}. Since sEMG measurements only require electrodes positioned on the forearm skin, this technique is very promising for future non-invasive wearable prosthetic hand control system \cite{zheng_surface_2022}.
    
    Transformers, which are the state-of-the-art networks for sequence processing \cite{vaswani_attention_2017, lin_survey_2022}, can be very efficient to process sEMG signals \cite{burrello_bioformers_2022}. However, the self-attention mechanism \cite{vaswani_attention_2017} used in conventional transformers requires to wait for large time windows, which induces a delay preventing fast online processing of continuous signals. Moreover, memory and computation of the self-attention mechanism scales quadratically with the sequence length.  
    
     In contrast, Recurrent Neural Networks (RNNs) integrate the concept of time into their operating model and are thus suited for continuous signals online processing.  Spiking Neural Networks (SNNs) \cite{zenke_superspike_2018, tavanaei_deep_2019} are a bio-inspired type of RNNs. They are very promising for low power applications because their neurons only transmit information when their membrane potential (an internal state of each neuron) reaches a threshold, and these events happen sparsely in time \cite{tavanaei_deep_2019}. Many research focus on building new hardware that leverage the inherent temporal sparsity of SNNs \cite{merolla_million_2014, liu_memory-efficient_2018, orchard_efficient_2021, pehle_brainscales-2_2022}. 

  In this paper, we propose an online transformer that makes use of a linearized sliding window attention mechanism \cite{beltagy_longformer_2020}. We adapt this attention mechanism for online processing of continuous signals by making it forward in time and serialized. Our online transformer thus performs inference for each token as they are generated. In order to leverage information from past inputs, we store information in the keys and the values of the attention mechanism, and we update this memory dynamically as the tokens are generated. The length of the sequences stored in the keys and the values is a hyper-parameter that we can tune to change the temporal depth of the information used in the attention mechanism, as well as the computational complexity and the memory usage.
  
    We test our model on a finger position regression through sEMG signals using the Non-Invasive Adaptive Hand Prosthetics Database 8 (NinaProDB8) dataset. First, we show that our online transformer allows users to process sEMG signals with high accuracy using solely very short time windows of 3.5 ms, which permits a very fine granularity in time of prosthetic hand control. Secondly, we show that selecting the temporal depth of the attention improves the results of signal processing and makes our model outperform a self-attention-based transformer, as well as previous state-of-the-art models. Finally, we show how our custom online attention mechanism allows us to SNNs inside the transformer architecture to increase the network sparsity, which in turn results in a reduction of the required number of synaptic operations by a factor of $\times5.3$ without loss of accuracy.

\section{Related work}
\paragraph{Deep Learning for Surface Electromyography processing}
    Although sEMG signals and muscle activity are correlated, their relation is unknown and processing sEMG signals remains very challenging because of electrical noise (e.g., interference, ground noise, crosstalk between electrodes), inter-subject variability (e.g., different forearm circumferences, muscle characteristics), and intra-subject variability (e.g., variation of the electrodes position or the skin conductivity from one day to the next) \cite{krasoulis_effect_2019}.
    
    Deep learning methods can leverage large datasets to extract the more relevant features despite noise or variability \cite{jaramillo-yanez_real-time_2020}. They can thus outperform conventional machine learning techniques like Support Vector Machine (SVM) \cite{milosevic_exploring_2018}. Moreover, deep networks can process raw sEMG signals whereas conventional networks require prior pre-processing like Principal Component Analysis \cite{phukpattaranont_evaluation_2018}, Linear Discriminant Analysis (LDA) \cite{phukpattaranont_evaluation_2018}, Fourier transforms \cite{taghizadeh_finger_2021}, and others.
    
    Deep learning has already been applied to sEMG signals processing using Temporal Convolutions \cite{tsinganos_improved_2019,zanghieri_robust_2020,zanghieri_semg-based_2021} and Recurrent Neural Networks (RNNs) \cite{anam_estimation_2020, koch_regression_2020,ilyas_evaluation_2022, li_approach_2022}. While the  ability to compute on the edge with restricted memory capacity and low power consumption are essential to the deployment of autonomous wearable prosthetic hand control systems, most deep learning techniques are computationally intensive. \citet{mukhopadhyay_classification_2018} have shown that the inherent sparsity of SNNs can be leveraged to reduce drastically the computational intensity of sEMG signals processing. \citet{burrello_bioformers_2022}  have shown that a transformer network can process sEMG signals with a limited memory usage and reduced number of Multiply-And-Accumulate (MAC) operations.
    
\paragraph{Transformers}
    Unlike RNNs, Transformers do not suffer from the vanishing gradient problem for learning features in time \cite{lin_survey_2022}, they do not have inductive biases made from assumptions about the data structure, and they can be trained very fast on GPUs since they can process an entire temporal sequence in parallel. The workhorse of Transformers is the self-attention mechanism, an operation that allows all the elements of a sequence to be compared with each other’s. For Natural Language Processing (NLP), the strength of self-attention is that is allows one token to be compared with present, past, and future tokens \cite{vaswani_attention_2017}. However, depending on the application,  conventional self-attention is not always the best choice. It has been shown that using local attention and sliding windows attention can lead to better results for long sequences in NLP \cite{beltagy_longformer_2020} and in Machine Vision \cite{hassani_dilated_2022}.

\paragraph{Transformers with Spiking Neural Networks.}
    SNNs, which mimic biological neural networks, are very promising for low power applications because their neurons only transmit information when their membrane potential (an internal state of each neuron) reaches a threshold, and these events happen sparsely in time \cite{tavanaei_deep_2019}. Integrating SNNs in a transformer architecture is challenging and not intuitive. Just as RNNs, SNNs have a temporal dynamic. Thus, each element of a sequence must be fed to RNNs or SNNs sequentially. In contrast, since the self-attention mechanism compares all the different elements of a sequence in parallel, Transformers require to wait for the completion of a sequence before computing. For instance, the transformer used in \cite{burrello_bioformers_2022} for sEMG classification used time windows of 150 ms. Naively stacking conventional self-attention layers and recurrent layers would then lead to undesirable delays due to the alteration between waiting time windows and processing sequences sequentially.
    
    \citet{yao_temporal-wise_2021} have used a type of attention mechanism to select the importance of event frames, and then process the events with a SNN. \citet{sabater_event_2022} have shown that a transformer can be used to process event-based vision sensor data more efficiently and accurately than convolutional neural networks.
    \citet{zhou_spikformer_2022} have used binarized self-attention to integrate sparsity in Transformers. 
    \citet{li_spikeformer_2022} have used a SNN as a pre-processing step for a transformer. It was also shown by \citet{gehrig_recurrent_2022} that Long-Term Short-Term (LSTM) units can be integrated inside a transformer architecture, but in this work the attention mechanism was spatial and not temporal. Finally, \citet{zhu_spikegpt_2023} have integrated spiking neurons inside a transformer architecture, but by using a custom attention mechanism that cannot be computed online.   
    
    In this paper, we introduce a transformer model that can perform attention in time online, and is compatible with spiking neurons at every layer of the architecture.     
    
\section{Methods}
\subsection{NinaproDB8: A Finger Position Regression Dataset}
    \begin{figure}
        \includegraphics[width=\linewidth]{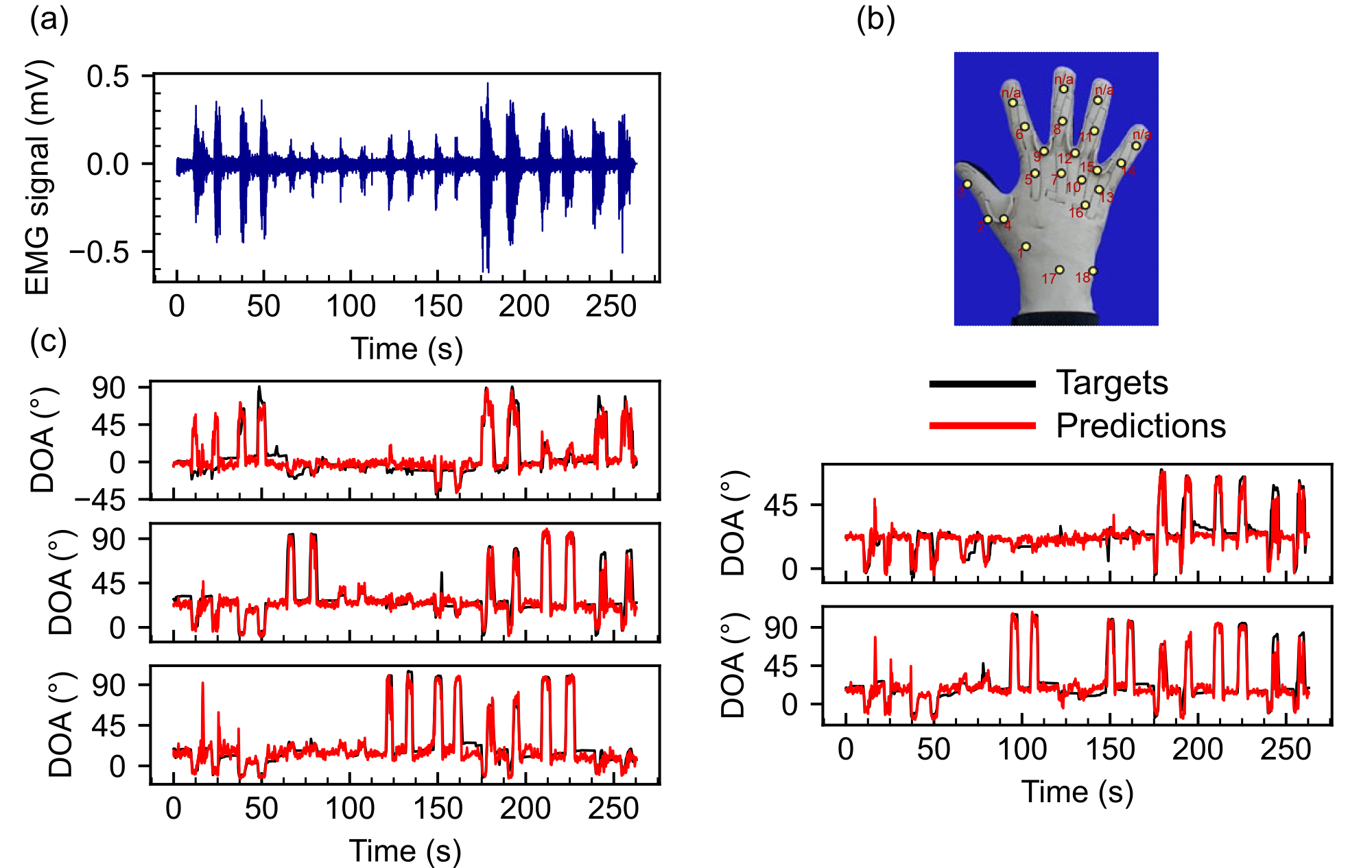}
        \caption{(a) Surface electromyography signal acquired using a 16 channel Delsys Trigno IM Wireless EMG system (see \citet{krasoulis_effect_2019}). The signal of only one out of the 16 channels is plotted. (b) The Cyberglove II is used for the acquisition of the ground truth finger-joint angles \cite{pizzolato_comparison_2017}. (c) Ground truth finger-joint angles and reconstruction with our Online Transformer model.}
        \label{fig:emg}
    \end{figure}
    In this work, we used the Non-Invasive Adaptive Hand Prosthetics Database 8 (NinaProDB8) \cite{krasoulis_effect_2019}, a public sEMG database made as a benchmark for estimation of kinematic finger position. Many deep learning efforts applied to sEMG focus on simple functional movement classification \cite{tsinganos_improved_2019, burrello_bioformers_2022, zheng_surface_2022,zanghieri_robust_2020}. However, sequence-to-sequence regression of finger position can lead to a wider range of gestures and can be more easily coupled to sensory feedback from robotic hands for a closed-loop precise control \cite{markovic_myocontrol_2018}. 
    
    The measurements of the database were made on 10 able-bodied subjects and two right trans-radial amputees. The sEMG signal, that is the input of our neural network (see Fig. \ref{fig:emg} (a)), are recorded using 16 electrodes (Delsys Trigno IM Wireless EMG system) positioned around the right forearm of the participants. The finger positions were measured using a dataglove, the Cyberglove II, 18-Degrees of Freedom (DoF) model, that measures the finger-joint angles that correspond to the dots in Fig. \ref{fig:emg} (b). The sEMG signals and the dataglove signals were up sampled to 2 kHz and post-synchronized. The details of the dataset can be found in \cite{krasoulis_effect_2019}.
    
    In order to disregard the irrelevant degrees of freedom and focus directly on motions relevant for prosthetic hand control, it has been shown by \citet{krasoulis_effect_2019} that we can convert the 18-DoF recorded by the dataglove into 5-Degrees of Actuation (DoA) using a simple linear transformation. The matrix used for this linear transformation can be found in the supplementary materials of \citet{krasoulis_effect_2019}. We used the DoA as targets of our neural network.
    
    Three datasets were recorded for each participant: the first two datasets (acquisition 1 and 2) comprised 10 repetitions of each movement and the third dataset (acquisition 3) comprised only 2 repetitions. We used both acquisition 1 and 2 as training set and acquisition 3 as testing set. In Fig. \ref{fig:emg} (a) and (c) we show the example of the testing set for subject 1 (target). 
    
    To facilitate the training of our neural network, we normalize each set of repetition by subtracting the sEMG signals by their mean and dividing by their standard deviation.

\subsection{Online Inference with a Custom Attention Mechanism}
\label{sec:online_inference}
    \begin{figure}
        \centering
        \includegraphics[width=\linewidth]{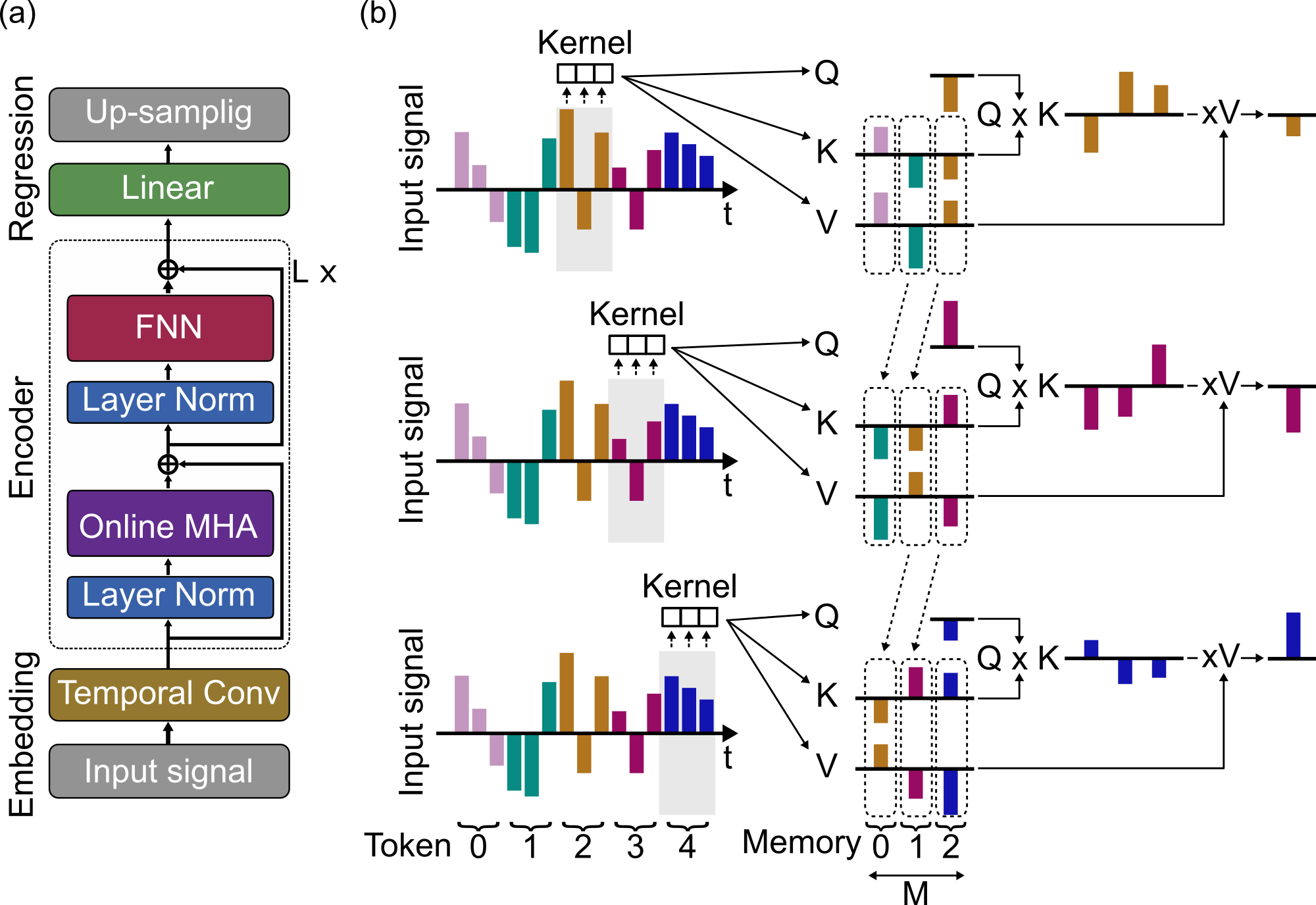}
        \caption{(a) Online Transformer neural network architecture. (b) Online Attention sketch: The different tokens are created by a temporal convolution (with a kernel size 3 and a stride 2 in this example). The tokens are linearly projected toward the queries, the keys and the values ($Q$, $K$, $V$). $Q$ matches only the present token whereas $K$ and $V$ store multiple previous tokens. The length $M$ of this memory is 3 in this example. At each time step, $K$ and $V$ forget the projection of the oldest token and store the projection of the new one. The mathematical operations of the online attention mechanism are described in section \ref{sec:online_inference}.}
        \label{fig:pipeline}
    \end{figure}
    In conventional transformers \cite{vaswani_attention_2017}, the entire self-attention stage is calculated in parallel. The elements of the input sequence of a self-attention layer are called tokens, and the operation of self-attention is described as
    \begin{equation}\label{eq:self_attention}
        \mathrm{Attention}(Q,K,V)=\mathrm{softmax}\left(\frac{Q\circ K^T}{\sqrt d}\right)\circ V \\
    \end{equation}
    where $\circ$ is the dot-product operator, $Q\in\mathbb{R}^{N\times d}$, $K\in\mathbb{R}^{N\times d}$, and $V\in\mathbb{R}^{N\times d}$ are respectively called the queries, the keys, and the values and are three different projections of the same sequence of tokens:
    \begin{subequations}
    \begin{align}
        Q&=W_{Q}x \label{eq:q_proj} \\
        K&=W_{K}x \label{eq:k_proj} \\
        V&=W_{V}x. \label{eq:v_proj}
    \end{align}
    \end{subequations}
    The attention dimension $d$ is the size of each token projection and $N$ is the sequence length. $W_{Q}\in\mathbb{R}^{d\times D}$, $W_{K}\in\mathbb{R}^{d\times D}$, and $W_{V}\in\mathbb{R}^{d\times D}$ are learnable weights matrices with with $D$ the embedding dimension, and $x\in\mathbb{R}^{N\times D}$ the input of the attention mechanism. 
    
    In the case of continuous signals (such as bio-medical signals), it is possible to split the input signal into finite time windows, and to wait for the end of each time window before carrying-out the inference (as in \citet{burrello_bioformers_2022}). However, this method induces delays due to waiting for the end of the time windows. Our online transformer uses a custom attention mechanism that can be computed online for each element of the sequence without delays. 
    
    To avoid waiting for future tokens, the tokens of time step $t_0$ are not compared with future tokens of time steps $t>t_0$. The information from previous tokens is stored in the keys and the values $K \in\mathbb{R}^{M\times d}$ and $V \in\mathbb{R}^{M\times d}$. Unlike for self-attention, here the size of $K$ and $V$ does not depend on the full sequence length, but solely on $M$, which is the number past time steps we choose to store.
    
\begin{figure}     
    \begin{minipage}{.5\linewidth}
        \begin{algorithm}[H]
            \caption{Inference with Online Attention}
            \begin{algorithmic}
            \State $K=\mathbf{0}$ \Comment{$K\in\mathbb{R}^{M\times d}$}
            \State $V=\mathbf{0}$ \Comment{$V\in\mathbb{R}^{M\times d}$}
            \State $t=0$
            \State $i=0$
            \Repeat 
                \State ${Q_t} \gets W_{Q}x_{t}$ \Comment{$Q_{t}\in\mathbb{R}^{d}$}
                \State ${K}_{i} \gets W_{K}x_{t}$\Comment{$K_i\in\mathbb{R}^{d}$}
                \State ${V}_{i} \gets W_{V}x_{t}$\Comment{$V_i\in\mathbb{R}^{d}$}
                \State $\mathrm{Attention}_{t} \gets \mathrm{softmax}\left(\frac{Q_{t} \circ K}{\sqrt{d}}\right) \circ V$
                \State $t \gets t+1$ \Comment{New time step}
                \State $i \gets i+1$ 
                \State $i \gets i\;(\bmod{M})$ 
            \Until{end of sequence}
            \end{algorithmic}
        \label{alg:online_attention}
        \end{algorithm}
    \end{minipage}
    \hspace{1em}
    \begin{minipage}{.4\linewidth}
      \includegraphics{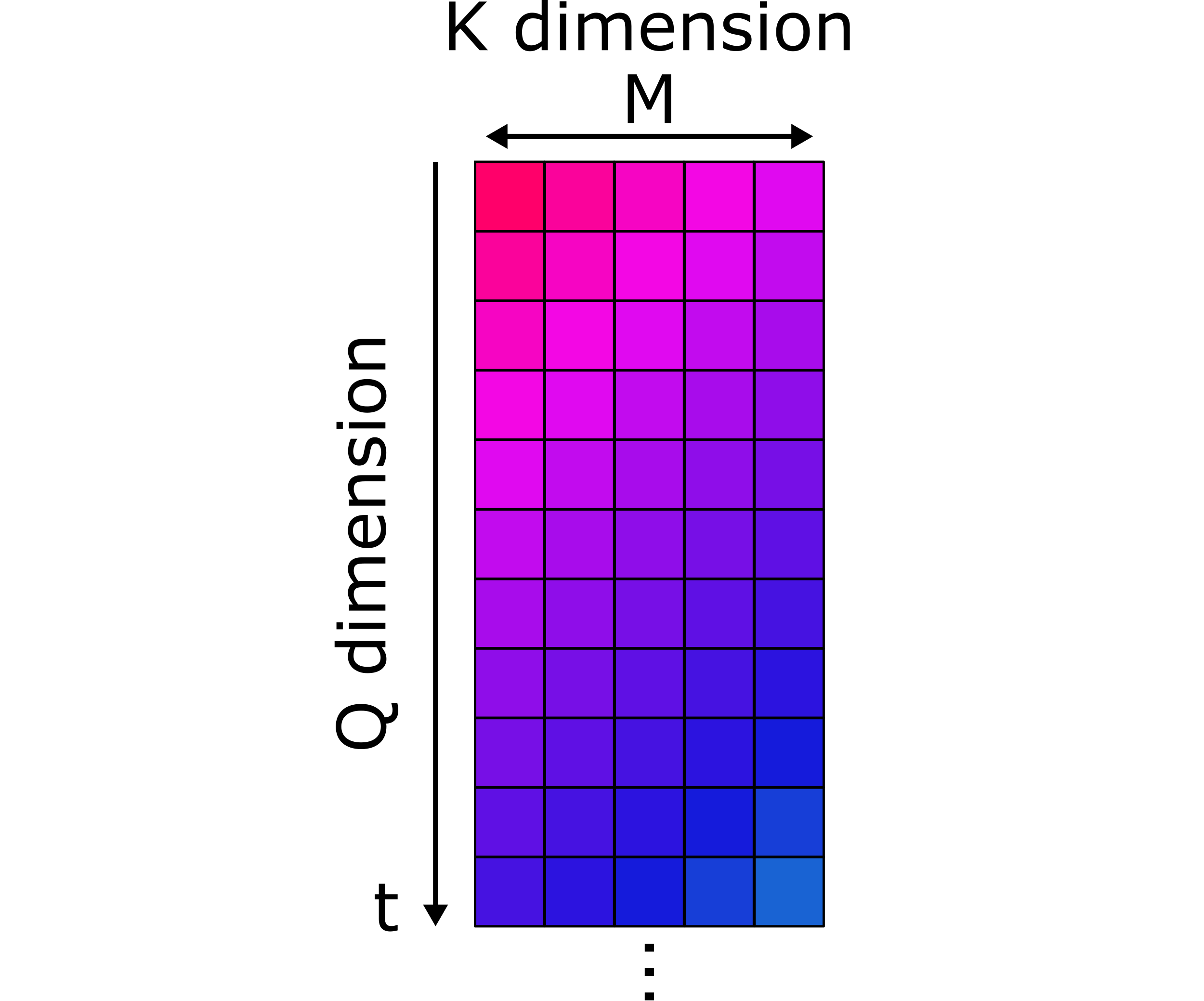}
        \caption{Sliding window attention. The same $K$ values are represented by the same colors.}
        \label{fig:sliding_attention}
    \end{minipage}
    \end{figure}

    $K$ and $V$ are initially zeroed. Then, the elements $K_{i} \in\mathbb{R}^{d}$ and $V_{i} \in\mathbb{R}^{d}$ are iteratively replaced token-wise using the projections of Eqs. \ref{eq:k_proj} and \ref{eq:v_proj}. At each time step, a single query $Q_{t} \in\mathbb{R}^{d}$ is also computed with Eq. \ref{eq:q_proj}, and the attention is computed as 
    \begin{equation}\label{eq:online_attention}
        \mathrm{Attention}_{t}=\mathrm{softmax}\left(\frac{Q_{t} \circ K}{\sqrt{d}}\right) \circ V
    \end{equation}
    The softmax is computed on the memory length dimension $M$. Since one different element of $K$ and $V$ is updated at each time step, and since the length of $K$ and $V$ is $M$, all their elements are updated with a frequency $\frac{1}{M}$. The above-mentioned procedure is summarized as the Algorithm \ref{alg:online_attention}.
    
    In contrast to the quadratic dependence with respect to sequence length for conventional self-attention ($O\left(N^2\right)$), the computational complexity of the sliding window attention mechanism is linear ($O\left(M N\right)$).

    Eq. \ref{eq:online_attention} shows that our attention mechanism can be computed time step wise instead of waiting for the end of large time windows to compute the attention in parallel. Now, we will show how this attention mechanism fits in our full neural network.

    \subsection{Neural Network Architecture}
    Our neural network consists of three blocks as depicted in Fig. \ref{fig:pipeline} (a): an embedding block that converts the raw EMG signal into a sequence of tokens, an encoder block that uses attention to find correlation between sequence elements, and a regression block that converts the output into five degrees of actuation. 

    The embedding is made of a temporal convolution layer. Convolutional layers have overlaps between input time windows, which means that unlike linear layers, they have an intrinsic order, and thus do not require positional embeddings \cite{vaswani_attention_2017}. The convolutional layer has $C=16$ input channels matching the 16 electrodes, and $D=64$ output channels. The network is tested with kernels of various sizes to vary the length of the input time window that matches one token. We chose to make an overlap of two time steps between time windows, which makes the convolutional layer stride be $s=k-2$, where $k$ is the kernel size. With a padding $p=1$, the number of tokens generated thus depends on the stride as 
    \begin{equation}\label{eq:n_tokens}
        N=\lfloor\frac{N_{\mathrm{samples}}}{s}\rfloor \\
    \end{equation}
    where $N_{\mathrm{samples}}$ is the number of processed samples of the input signal.
    
    The encoder is described as 
    \begin{equation}\label{eq:encoder}
    	\begin{array}{cll}
        	f\left(x\right)&=&x+\mathrm{MHA}\left(\mathrm{LN}\left(x\right)\right)\\z\left(f\left(x\right)\right)&=&f\left(x\right)+\mathrm{FNN}\left(\mathrm{LN}\left(f\left(x\right)\right)\right)\\
    	\end{array}
	\end{equation}
    where $\mathrm{LN}$ is a layer norm layer. $MHA$ is the multi-head attention layer with $h=8$ heads computed in parallel using Eq. \ref{eq:online_attention}, with an attention dimension $d=32$. After that attention is computed, the $h$ heads are concatenated and projected into dimension $D=64$. $FNN$ is a Feedforward Neural Network with one hidden layer of 128 GeLU units \cite{hendrycks_gaussian_2020} and a dropout layer with probability 0.2. The linear projections of $\mathrm{FNN}$ are applied token-wise so that that they can be computed online. The entire encoder block can be repeated and stacked $L$ times, but here we chose to keep $L=1$. The backbone of our neural network is inspired from \citet{burrello_bioformers_2022}. 
    
    Finally, the regression block consists of a linear layer that projects each token from a dimension $D = 64$ to a dimension 5 (the number of degrees of actuation we perform the regression on), and an up sampling layer that duplicates the output of each token to generate as many samples as there are in the target signal (an example of target signal is shown in Fig. \ref{fig:emg} (c)). The up sampling factor is equal to the stride that we use in the convolutional embedding layer (see Eq. \ref{eq:n_tokens}). 
    
    \subsection{Increasing the network sparsity with binarization and spiking neurons}
    
    In order to reduce the number of required operations, we increase the network sparsity by using binarization and Leaky Integrate and Fire (LIF) units \cite{zenke_superspike_2018,tavanaei_deep_2019} units. We test two sparse models. In the first one, we binarize the output of the convolutional embedding, we binarize the projections $Q$, $K$, and $V$, and we replace the FNN by a SNN with a first layer of 128 LIF units, and a second of $D=64$ LIF units. The second sparse model is similar to the first one, but instead of binarizing $Q$, $K$, and $V$, we replace each projection of Eqs. \ref{eq:q_proj}, \ref{eq:k_proj}, and  \ref{eq:v_proj} by a single spiking layer of $d=32$ LIF units, which adds an additional dynamic to the model.
    
    Binarization is done by applying a Heaviside function.
    The dynamics of the LIF units are defined by
    \begin{subequations}
        \begin{align}            U_t&=\alpha\left(1-S_{t-1}\right)U_{t-1}+\left(1-\alpha\right)I_{t-1} \label{eq:U} \\
            I_t&=\beta I_{t-1}+\left(1-\beta\right)Wx_t \label{eq:I} \\
            S_t&=H\left(U_{t-1}-\mathrm{\Theta}\right) \label{eq:S}
        \end{align}
    \end{subequations}
    where $t$ is the index of the tokens, $U$ is the membrane potential, $I$ is the synaptic current, S is the spike response, H is the Heaviside function, $\alpha=0.95$, $\beta=0.9$, and $\mathbf{\Theta}=1$. The outputs of the $Q$, $K$, and $V$ projections are the spike responses $S_t$ (see Eq. \ref{eq:S}). The outputs of the first layer of the SNN replacing the FNN are the spike responses $S_t$, and the outputs of the second layer are the membrane potentials $U_t$ (see Eq. \ref{eq:U}).
    
    Because the Heaviside function is not differentiable, during training the gradient of the different Heaviside functions (used for binarization and LIF units) are replaced by the SuperSpike surrogate gradient \cite{zenke_superspike_2018, neftci_surrogate_2019}. To preserve the sparsity between the embedding and the encoder block, we remove the layer norm layer that precedes the embedding when the embedding is binarized. In addition, we remove the dropout layers in the two sparse models. The softmax of the attention mechanism is only computed on non-zero elements.
    
    \subsection{Training}
    To speed up training, the attention block is computed in parallel. Projections  $Q\in\mathbb{R}^{N\times d}$, $K\in\mathbb{R}^{N\times d}$, and $V\in\mathbb{R}^{N\times d}$ are computed for an entire time window with $N$ tokens. The keys and values are then unfolded into sliding windows of size $M$ and stride 1, similarly as for a convolution (see in Fig. \ref{fig:sliding_attention} an example of sliding window attention). The product between queries and keys is thus computed as 
    \begin{equation}\label{eq:attention_training_matrix}
            \left[\begin{matrix}
            Q_{0}K_{1-M} & \cdots & Q_{0}K_{-2} & Q_{0}K_{-1} & Q_{0}K_{0} \\
            Q_{1}K_{2-M} & \cdots & Q_{1}K_{-1} & Q_{1}K_{0} & Q_{1}K_{1} \\
            Q_{2}K_{3-M} & \cdots & Q_{2}K_{0} & Q_{2}K_{1} & Q_{2}K_{2} \\
            \vdots & \iddots & \vdots & \vdots & \vdots \\
            Q_{N}K_{N-M} & \cdots & Q_{N-1}K_{N-3} & Q_{N-1}K_{N-2} & Q_{N-1}K_{N-1} \\
            \end{matrix}\right].
    \end{equation}
    Since the keys $K_{i<0}$ are forbidden values, we mask them by replacing them with $-\infty$ as in \citet{vaswani_attention_2017}, so that they are not computed in the softmax (see Eq. \ref{eq:online_attention}). The values $V_{i<0}$ are simply zeroed.

    To improve training, we developed a simple data augmentation protocol: first, the training set signals are sliced into time windows of $N_\mathrm{samples} = 2000$ samples (which corresponds to $1\,$s since the sampling rate is $2\,$kHz). Then, each time window is duplicated 64 times. For data augmentation, the beginning of each of this duplicated time window is shifted with a random number sampled in a uniform distribution between 0 and 2000. Finally, the resulting time windows are shuffled to create the training dataset.
    
    We trained each network for each subject for 10 epochs using the Adam optimizer \cite{kingma_adam_2017}, a learning rate of ${10}^{-3}$, batch sizes of 64, and since the metric we want to minimize is the mean average error (MAE) over the 5 degrees of freedom (DoA), we used the L1 loss function.
    
    For the sparse models, we added a sparsity loss function term \cite{yan_backpropagation_2022} to the global loss to increase the sparsity of the embedding, the queries, keys and values such that the total loss is:
    \begin{equation}\label{eq:sparsity_loss}
        \mathcal{L}\left(y,\hat{y}\right)=\parallel y_{i,j}-\hat{y}_{i,j} \parallel_1
        -\frac{1}{2}\lambda\left(\parallel x \parallel_{2}+
        \parallel \mathrm{Concat}\left(Q,K,V\right) \parallel_{2}\right)
    \end{equation}
    with $y$ the network outputs, $\hat{y}$ the targets, $x$ the embeddings and $\lambda=1$.

    In this study, we simply trained and tested datasets independently for each subject. To improve accuracy and repeatability in future studies, it is also possible to use transfer learning: the network can learn from multiple subjects before fine-tuning and testing on a new subject, as in \cite{lehmler_deep_2022}.

\section{Results}
    
\begin{figure}
        \centering
        \includegraphics[width=\linewidth]{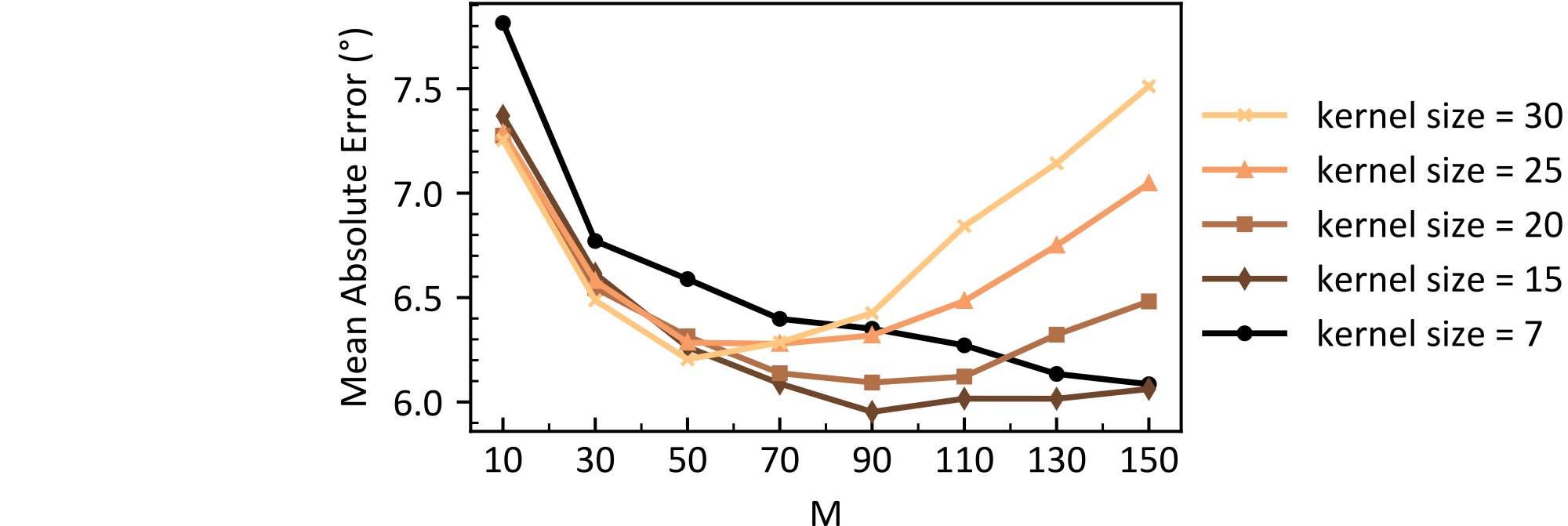}
        \caption{Mean absolute error averaged over the 12 subjects versus the number of stored tokens $M$ for kernel embedding sizes of 7, 15, 20, 25, and 30 (from black to light yellow curves).}
        \label{fig:svs_sweep}
    \end{figure}

    In Fig. \ref{fig:emg} (c) we show an example of the regression results for a sparse online transformer with a embedding convolution kernel size $k = 7$ and a memory length $M = 150$.
    We first investigate how $k$ and $M$ affect the final accuracy. For this study, we use the network without sparsity. Since $s=k-2$, we simultaneously change $s$ and $k$, and thus the number of tokens $N$ generated for a given time window (see Eq. \ref{eq:n_tokens}). The memory length $M$ defines how many past tokens are used in the  attention mechanism. The time length of the signal used to store information in $K$ and $V$ is thus:
    \begin{equation}\label{eq:t_memory}
        \tau_{\mathrm{memory}}=\frac{M\times s}{\mathrm{SamplingRate}}
        .
    \end{equation}
    In Fig. \ref{fig:svs_sweep} we plot the mean absolute error (MAE) over the different degrees of actuation for values of $M$ swept between 10 and 150 with intervals of 20, and for five different values $k=$ 7, 15, 20, 25, and 30 (which correspond to $s=$ 5, 13, 18, 23 and 28). While sweeping $M$, we see for $k=$ 15, 20, 25, and 30 that the MAE reaches a minimum and then increases. It shows that there is then an optimum value of memory length for each kernel size, and we see that this optimum value decreases with the kernel size and thus with the stride. Using Eq. \ref{eq:t_memory}, this result indicates that there is an optimum length of information $\tau_\mathrm{memory}$ used in the attention mechanism, and that past that point increasing the stored information does not increase the accuracy. 
    Then, we compare our different models using each time a kernel size $k = 7$ and a memory length is $M=150$. As we see in Fig \ref{fig:svs_sweep}, these parameters lead to the best accuracy using the shortest time window for each token. For this study we also measure the 10°-accuracy and the 15°-accuracy, which are respectively the proportion of time samples that lead to mean average errors inferior to ten and fifteen degrees \cite{koch_regression_2020}. These additional metrics are important to measure the accuracy of the prediction within a margin of error. The different results are shown in Table \ref{table:ninaprodb8_results}. The mean and standard deviation of each metric are computed over the 12 subjects of the NinaproDB8 dataset.
    
    To see the impact on  our online transformer with custom sliding window attention mechanism, we compare it to a conventional Transformer with self-attention. For the three metrics, our online transformers outperform the transformer with conventional self-attention (see Table \ref{table:ninaprodb8_results}). This results further reveals the importance of selecting relevant information, and that for sEMG signal processing, it is likely more important to use local information from the past than global information from both past and future. 

    \begin{table}\label{table:ninaprodb8_results}
      \caption{Results of regression on the Ninapro DB8 database with different models}
      \centering
      {\small  
      \begin{tabular}{llllll}
        \toprule
        \textbf{Model} & \textbf{MAE (°)}$\downarrow$ & \textbf{10°-accuracy}$\uparrow$ & \textbf{15°-accuracy}$\uparrow$ & $\mathbf{\tau_{min}}\downarrow$ & \textbf{MMAC ops}$\downarrow$\\&&&&&\textbf{/inference}\\
        \toprule
        SVM$^{\ast}$ \cite{zanghieri_semg-based_2021} & 7.28 & 0.79 & 0.88 & 60 ms & --\\
        \midrule
        MLP$^{\ast}$ \cite{zanghieri_semg-based_2021} & 7.14 & 0.80 & 0.89 & 60 ms &  --\\
        \midrule
        TempConv \cite{zanghieri_semg-based_2021} & 6.89 & 0.81 & 0.90 & 128 ms & 3.2 \\
        \midrule
        LSTM \cite{koch_regression_2020} & 7.04 &  --&  --& 10 ms &  --\\
        \midrule
        Transformer$^{\dagger}$ & 6.62$\pm$1.52 & 0.87$\pm$0.07 & 0.92$\pm$0.05 & 2 s & 3,769 \\
        \midrule
        Online Transformer$^{\dagger}$ & \textbf{6.10$\pm$1.50} & \textbf{0.86$\pm$0.07} & \textbf{0.94$\pm$0.05} & \textbf{3.5 ms}& 5.3 \\
        \midrule
        Online Transformer with \\
        binary embedding and QKV, & \textbf{6.08$\pm$1.27} & \textbf{0.87$\pm$0.06} & \textbf{0.94$\pm$0.04} & \textbf{3.5 ms} & \textbf{1.4} \\ spiking FNN$^{\dagger}$ \\
        \midrule
        Online Transformer with \\
        binary embedding, & \textbf{6.16$\pm$1.39} & \textbf{0.87$\pm$0.07} & \textbf{0.94$\pm$0.04} & \textbf{3.5 ms} & \textbf{1.0} \\ spiking QKV and FNN$^{\dagger}$ \\
        \midrule
        $^{\ast}$ With prior feature extraction. & & & &\\
        $^{\dagger}$ This work.\\
        \bottomrule
      \end{tabular}}
    \end{table}
    Our two sparse models reach similar accuracy than our non-sparse online transformer (and thus also better accuracy than equivalent conventional transformer), and respectively lead to a reduction of the number of required Multiply-And-Accumulate operations (MAC ops) by factors of $3.8 \times$ and $5.3 \times$ compared to the non-sparse online transformer (the activation function operations are not included in these calculations). 
    The method used to compute the number of required operations is described in the Appendix.
    
    Moreover, we see that our three online transformer models outperform LSTMs \cite{koch_regression_2020} by at least 0.88° of MAE, outperform Temporal Convolutions \cite{zanghieri_semg-based_2021} with at least 0.76° of MAE (previous SoTA on NinaproDB8 dataset). To compare the inference speed of the different methods, we define the minimum time of computing as the length of the time windows used for each inference step, which for our online transformer is $ \tau_{\mathrm{min}}=\frac{k}{\mathrm{SamplingRate}}$,
    with $k$ the embedding convolution kernel size. Since $k=7$ and the sampling rate is 2 kHz, our network can compute with a minimum latency of 3.5 ms, which is shorter than any previous methods and in particular more than $30\times$ shorter than the Temporal Convolutional network \cite{zanghieri_semg-based_2021} which was the previous SoTA for the Ninapro DB8 dataset.
    
\section{Conclusion}
    In this work, we developed an online transformer model that leverages sliding window attention to process tokens one at the time. We have shown that the locality of the sliding window makes it more efficient than self-attention. The proposed method makes sEMG signal processing with very short time windows (3.5 ms) possible, and sets the new state-of-the-art on the prosthetic hand control NinaproDB8 dataset. 
    Using sliding window attention, our model also solves the problem of the integration of SNNs temporal dynamics in Transformers. 
    We used a combination of binarization and SNNs to increase the network sparsity, thus reducing the number of required operation up to a factor $5.3\times$. 
    In conclusion, this work is a step toward precise, smooth, and low-power Human-Machine Interfacing, and holds great promises for future neuromorphic transformer models. \blfootnote{Python codes available at https://github.com/NathanLeroux-git/OnlineTransformerWithSpikingNeurons}

\medskip


\appendix

\section{Appendix}

To compute the number of synaptic MAC operations of our different models, we used the following set of equations, that represent the different operations (activation function excluded) of our models:

\begin{subequations}
    \begin{align}  
        \mathrm{\#EmbeddingMACs}     &= k\times C \times D \times 32 \label{eq:shsb} \\
        \mathrm{\#QKVProjectionMACs} &= \left(1-\mathrm{EmbeddingSparsity}\right) \times 3 \times D \times d \times h \times 32 \\
        \mathrm{\#QKProductMACs}     &=\parallel QK \parallel_1 \times M \times h \times 32 \\
        \mathrm{\#VProductMACs}      &=\left(1-\mathrm{VSparsity}\right) \times d \times M \times h \times 32 \\
        \mathrm{\#ConcatMACs}        &=\left(1-\mathrm{AttentionSparsity}\right) \times d \times h \times D \times 32 \\
        \mathrm{\#TotalAttentionMACs}    &=\mathrm{\#QKVProjectionMac}+ \mathrm{\#QKProductMacs}  \\
        &                     +\mathrm{\#VProductMacs}+\mathrm{\#ConcatMACs} \nonumber \\    
        \mathrm{\#FFL1MACs}           &=D \times H \times 32 \\
        \mathrm{\#FFL2MACs}           &=\left(1-\mathrm{FFL1Sparsity}\right) \times H \times D \times 32 \\
        \mathrm{\#RegressionMACs}     &=D \times 5 \times 32  \\    
        \mathrm{\#TotalMACs}          &=\mathrm{\#EmbeddingMacs}+\mathrm{\#TotalAttentionMACs} \\
        &                     +\mathrm{\#FFL1MACs}+\mathrm{\#FFL2MACs}+\mathrm{\#RegressionMACs} \nonumber
    \end{align}
\end{subequations}

\end{document}